\documentclass[letterpaper]{article} 
\usepackage{aaai21}  
\usepackage{times}  
\usepackage{helvet} 
\usepackage{courier}  
\usepackage[hyphens]{url}  
\usepackage{graphicx} 
\usepackage{natbib}  
\usepackage{caption} 
\usepackage[utf8]{inputenc}
\usepackage{algorithm}
\usepackage{algpseudocode}
\usepackage{mathtools}
\usepackage{amssymb}
\usepackage{graphicx}

\usepackage{subfigure}

\begin{document}
\title{Successive Halving Top-k Operator}
\author {
    Michał Pietruszka\textsuperscript{\rm 1,2},
    Łukasz Borchmann\textsuperscript{\rm 1,3},
    Filip Graliński\textsuperscript{\rm 1,4}
    \\
}
\affiliations {
    \textsuperscript{\rm 1} Applica.ai, Poland \\
    \textsuperscript{\rm 2} Jagiellonian University, Poland \\
    \textsuperscript{\rm 3} Poznan University of Technology, Poland \\
    \textsuperscript{\rm 4} Adam Mickiewicz University, Poland \\
    \{michal.pietruszka, lukasz.borchmann\}@applica.ai
}

\date{September 2020}
\maketitle
\begin{abstract}
\begin{quote}
We propose a differentiable successive halving method of relaxing the top-k operator, rendering gradient-based optimization possible. The need to perform softmax iteratively on the entire vector of scores is avoided by using a tournament-style selection. As a result, a much better approximation of top-k with lower computational cost is achieved compared to the previous approach.
\end{quote}
\end{abstract}

\section{Problem Statement}
Let $n$ denote the number of $d$-dimensional vector representations, resulting in a matrix $\mathnormal{E \in \mathbb{R}^{n\times d}}$. A scalar $v_i$ is assigned to each representation $\mathnormal{E_i \in \mathbb{R}^d}$. We want to select $k$ vectors out of $n$ in $E$, so that the ones achieving the highest scores in $v$ will be passed to the next layer of a neural network. The soft top-k operator $\mathnormal{\Gamma}$ is defined for this task, such that $\mathnormal{\Gamma\colon \mathbb{R}^{n \times d} \times  \mathbb{R}^{n} \rightarrow  \mathbb{R}^{k \times d}}$ is selecting $\mathnormal{k}$ representations which will form the input to the next neural network layer. The operation has to be differentiable w.r.t. $v$.

\section{Previous Approaches}
The previously introduced solution, presented in Algorithm~\ref{algo-iter-topk} was based on an approximation of top-k selection with iterative softmaxes \cite{goyal2017continuous} and has a complexity of $\mathcal{O}(n)$. In each of the $k$ steps of the algorithm, the weight distribution $p_i$ is calculated through a peaked-softmax function on a modified vector of scores. 
The obtained $p$ values are then used to perform a linear combination of all representations in $E$, leading to an approximated $TopK(E)$ matrix. At the end of each step, the score which is highest at the moment is overwritten by some large negative number that guarantees future weights to be negligible.

\begin{algorithm}[t]

    \caption{Iterative Top-K Selection \cite{goyal2017continuous}}
    \begin{algorithmic}[1]
    \Procedure{ContinuousTopK}{$E, v$}
    \State $E' \gets 0_{k,d}$
    \For{$i \gets 1, k$}
    \State $m \gets \Call{Max}{v}$
    \State $a \gets \Call{Argmax}{v}$
    \State $p_i \gets \Call{PeakedSoftmax}{-(v-m\cdot \textbf{1})^2}$ 
    \State $E'_i \gets p_i \cdot E$
    \State $v_{a} \gets - 10000$ \Comment{Masking out max.}
    \EndFor
    \State \textbf{return} $E'$
    \EndProcedure\\
    
    \Procedure{PeakedSoftmax}{$v$}
    \State $v' \gets 0_{n,1}$
    \State $denom = \sum_j{\operatorname{exp}(v_j)}-\operatorname{exp}(\Call{Max}{v})$
    \For{$i \gets 1, n$}
    \State $v'_i \gets \operatorname{exp}(v_i) \cdot \frac{1}{denom}$
    \EndFor
    \State \textbf{return} $v'$ \Comment{Probability distribution.}
    \EndProcedure
    \end{algorithmic}
    \label{algo-iter-topk}
\end{algorithm}

\section{Novel Successive Halving Top-k}

Algorithm~\ref{algo-soft-topk} presents the Successive Halving Top-K selection mechanism we propose. In short, we perform a tournament soft selection, where candidate vectors are compared in pairs $(i, j)$, until only $k$ remained. After each round of the tournament, a new $E'$ and $v'$ are composed as a linear combination of these pairs with weights based on their respective scores. Each new vector is calculated as:
\begin{equation*}
    E_{i} \cdot w_i + E_{j} \cdot w_j,
\end{equation*}
where the $w_i, w_j$ is the result of a boosted softmax over scores $v_i, v_j$.
Analogously, the new-round's scores are calculated as:
\begin{equation*}
    v_{i} \cdot w_i + v_{j} \cdot w_j.
\end{equation*}
Weights are calculated using the $\operatorname{BoostedSoftmax}$ function, increasing the pairwise difference in scores between $v_i$ and $v_j$. Here, multiple functions can be used. For example, softmax with base greater than $e$ or, equivalently, $\operatorname{softmax}(Cx,Cy)$ with constant $C \gg 1$.

One round halves the number of elements in $E$ and $v$. We perform it iteratively unless the size of $E$ and $v$ matches the chosen value of $k$, thus achieving $\mathcal{O}(\log_2(n/k))$ time-complexity. 

To improve convergence towards selecting the real top-k, it is desired to permute $v$ and $E$ first. In our Algorithm \ref{algo-soft-topk}, we sort vectors in $\mathnormal{E}$ by their scores $\mathnormal{v}$ and then make pairs is such a way that the $\mathnormal{i}$-th highest-scoring vector will be paired with the $\mathnormal{(n-i+1)}$-th highest-scoring vector, marked with the $j$-th index. Here, a simple non-differentiable sorting operation suffices. Note that the selection of preferable permutation itself makes the process only partially differentiable. In the case of modern CPUs, the cost of sorting is practically negligible.

\begin{algorithm}[t]

    \caption{Successive Halving Top-K Selection}
    \begin{algorithmic}[1]
    \Procedure{TopK}{$E, v$}
    \For{$i \gets 1, \log_2(\lceil n/k \rceil)$}
    \State $E, v \gets \Call{Sort}{E, v}$
    \State $E, v \gets \Call{Tournament}{E, v}$
    \EndFor
    \State \textbf{return} $E$
    \EndProcedure\\

    \Procedure{Sort}{$E, v$}
    \State $E' \gets (E_1,E_2,..), $ where $v_i \ge v_{i+1}$ and $v_i \in v$ 
    \State $v' \gets (v_1,v_2,..), $ where $v_i \ge v_{i+1}$ and $v_i \in v$
    \State \textbf{return} $E', v'$
    \EndProcedure\\
    
    \Procedure{Tournament}{$E, v$}
    \State $n  \gets \frac{1}{2} \lVert v \rVert$ \Comment{Target size}
    \State $d  \gets \lVert E_{*,1} \rVert$ \Comment{Representation depth}
    \State $v' \gets 0_{n,1}$  
    \State $E' \gets 0_{n,d}$
    \For{$i \gets 1, n$}
        \State $w \gets \Call{BoostedSoftmax}{v_i, v_{2n-i+1}}$
        \State $E'_i \gets E_i \cdot w_0 + E_{2n-i+1} \cdot w_1$
        \State $v'_i \gets v_i \cdot w_0 + v_{2n-i+1} \cdot w_1$
    \EndFor
    \State \textbf{return} $E', v'$
    \EndProcedure

    \end{algorithmic}
    \label{algo-soft-topk}
\end{algorithm}

\section{Evaluation}

\begin{figure}[t]
    \begin{subfigure}
    \centering
    \includegraphics[width=1.0\linewidth,trim={0.4cm 0.7cm 0.4cm 0.4cm},clip]{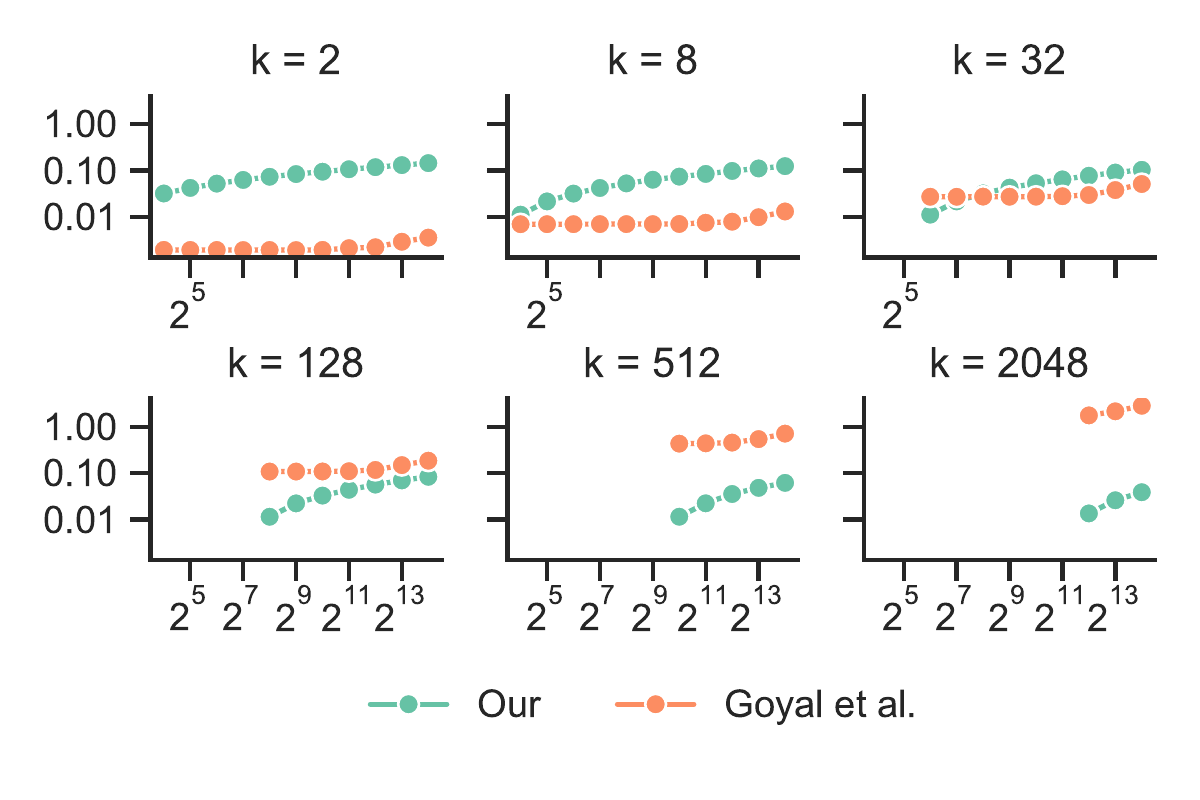}
    \caption{Number of seconds required to process a batch of sequences ($Y$-axis). Results depending on $n$ ($X$-axis). The lower the better.
    }
    \label{fig:performance}
    \end{subfigure}
    \vspace{0.8cm}
    \begin{subfigure}
    \centering
    \includegraphics[width=1.0\linewidth,trim={0.4cm 0.7cm 0.5cm 0.4cm},clip]{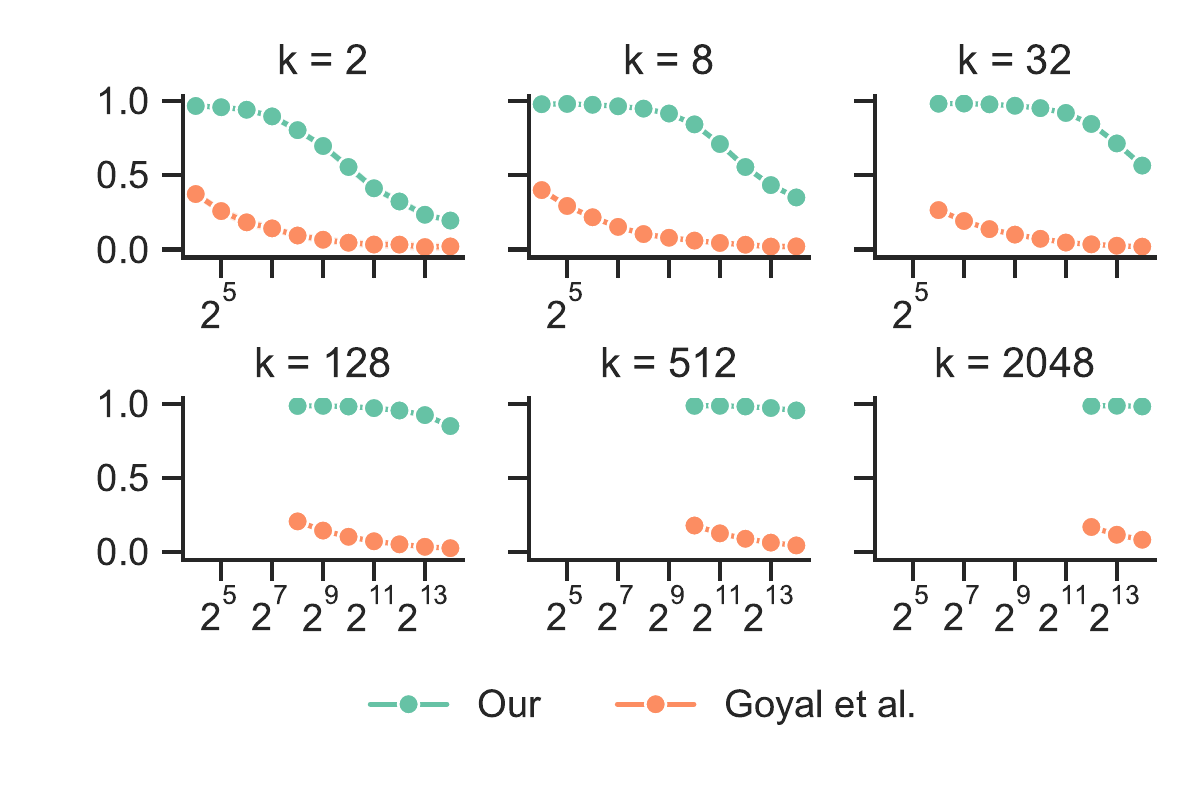}
    \caption{Approximation quality  ($Y$-axis) in the $nCCS$ metric. The higher the better.  
    }
    \label{fig:performance2}
    \end{subfigure}

\end{figure}

We assessed the performance of both algorithms experimentally, on randomly sampled matrices $E$ such that $\mathnormal{E\sim \mathcal{U}}[-1,1]$ and scores $\mathnormal{v\sim \mathcal{U}}[0,1]$. The selected $k$ top-scoring vectors were compared to the real top-k selection using normalized Chamfer Cosine Similarity (nCCS) as given:
\begin{equation*}
nCCS = \frac{1}{k} \sum_{i =1}^{k} \max_{j \in [1,k]}(\operatorname{cos}( y_i, \hat{y_j}))
\end{equation*}

Additionally, we measured an average time for processing a batch of size $16$ on the NVIDIA A100 GPU, for $n$ in range $2^{4,5,..,14}$ and $k$ in range $2^{1,2,..,11}$.

We addressed the question of how both algorithms differ in terms of speed (Figure \ref{fig:performance}) and quality (Figure \ref{fig:performance2}), depending on $k$ and $n$ choices.
One can notice that the higher the choice of $k$, the faster our algorithm is, and the slower is the iterative baseline of \cite{goyal2017continuous} as predicted by their complexities. Our solution's qualitative robustness is proven by achieving higher similarity to real top-k for any given $k$. The score degrades as the number of rounds in the tournament increases, as each round introduces additional noise. 

\section{Summary}
We proved that top-k could be relaxed using tournament-style soft-selection, leading to a better computational complexity and improved approximation quality than in the case of a previous solution. Furthermore, the Successive Halving Top-K we proposed performs robustly for large values of $k$ and $n$. The advance was achieved by limiting the number of elements softmax is performed on, as well as by a reduction in the number of steps required.

We expect our algorithm to perform better when employed as a neural network layer, due to a shorter chain of backpropagation's dependencies. We are excited about the future of differentiable top-k based selection and plan to apply them to downstream tasks.

\bibliography{bibliography}

\end{document}